\documentclass[10pt,twocolumn,letterpaper]{article}

\usepackage{cvpr}
\usepackage{times}
\usepackage{epsfig}
\usepackage{graphicx}
\usepackage{amsmath}
\usepackage{amssymb}

\usepackage{subfigure} 
\usepackage{xcolor}
\usepackage{csquotes}
\usepackage[numbers]{natbib}
\usepackage{algorithm}
\usepackage{algorithmic}
\usepackage{multirow}
\usepackage{tabularx}
\usepackage{array}
\usepackage[font=footnotesize]{caption}
\usepackage{capt-of}
\usepackage[english]{babel}
\usepackage{booktabs}

\usepackage{tikz}
\usetikzlibrary{positioning}

\newcommand{\E}{\mathbb{E}}
\newcommand{\Var}{\operatorname{Var}}
\newcommand{\real}{\operatorname{Re}}
\newcommand{\imag}{\operatorname{Im}}
\newcommand{\ci}{\mathrm{i}}
\newcommand{\id}{\mathrm{id}}
\newcommand{\Z}{\mathbb{Z}}
\newcommand{\R}{\mathbb{R}}
\newcommand{\C}{\mathbb{C}}
\newcommand{\equalCLT}{\overset{\hphantom{\text{(CLT)}}}{=}}
\newcommand{\supp}{\operatorname{supp}}

\newcommand{\myparagraph}[1]{\vspace*{1ex}\noindent\textbf{#1}}
\newcommand{\norm}[1]{\left\lVert #1 \right\rVert}

\newlength{\myx} 
\newlength{\myy} 
\newcommand\includegraphicstotab[2][\relax]{%
	\settowidth{\myx}{\includegraphics[{#1}]{#2}}%
	\settoheight{\myy}{\includegraphics[{#1}]{#2}}%
	\parbox[c][1.1\myy][c]{\myx}{%
	\includegraphics[{#1}]{#2}}%
}

\usepackage[pagebackref=true,breaklinks=true,letterpaper=true,colorlinks,bookmarks=false]{hyperref}

\cvprfinalcopy 

\ifcvprfinal\fi

\begin{document}

\title{Learning Steerable Filters for Rotation Equivariant CNNs}

\author{Maurice Weiler$^{1,2}$ \qquad Fred A. Hamprecht$^{2}$ \qquad Martin Storath$^{2}$\\
	$^1$AMLab / QUVA Lab, University of Amsterdam \qquad $^2$HCI/IWR, University of Heidelberg\\
	{\tt\small m.weiler@uva.nl \qquad  \{fred.hamprecht, martin.storath\}@iwr.uni-heidelberg.de}
}

\maketitle

\begin{abstract}
	In many machine learning tasks it is desirable that a model's prediction transforms in an equivariant way under transformations of its input.
	Convolutional neural networks (CNNs) implement translational equivariance by construction; for other transformations, however, they are compelled to learn the proper mapping.
	In this work, we develop \emph{Steerable Filter CNNs} (\mbox{SFCNNs}) which achieve joint equivariance under translations and rotations by design.
	The proposed architecture employs steerable filters to efficiently compute orientation dependent responses for many orientations without suffering interpolation artifacts from filter rotation.
	We utilize group convolutions which guarantee an equivariant mapping.
	In addition, we generalize He's weight initialization scheme to filters which are defined as a linear combination of a system of atomic filters.
	Numerical experiments show a substantial enhancement of the sample complexity with a growing number of sampled filter orientations and confirm that the network generalizes learned patterns over orientations.
	The proposed approach achieves state-of-the-art on the rotated MNIST benchmark and on the ISBI 2012 2D EM segmentation challenge.
\end{abstract}

\vspace{-1.0ex}
\section{Introduction}
\vspace*{-.75ex}
Convolutional neural networks are extremely successful predictive models when the input data has spatial structure.
One principal reason is that the convolution operation exhibits translational equivariance so that feature extraction is independent of the spatial position.
For many types of images it is desirable to make feature extraction orientation independent as well.
Typical examples are biomedical microscopy images or astronomical data which do not show a prevailing global orientation.
Consequently, the output of a network processing such data should be equivariant w.r.t.~the orientation of its input
-- if the input is rotated, the output should transform accordingly.
Even when there is a predominant direction in an image as a whole, the low level features in the first layers such as edges usually appear in all orientations; see e.g. the filterbanks visualized in \cite{DBLP:journals/corr/ZeilerF13}.
In both cases, conventional CNNs are compelled to learn rotated versions of the same filter, introducing redundant degrees of freedom and increasing the risk of overfitting.

\vspace{-.5ex}
\subsection{Contribution}
\vspace{-.75ex}
We propose a rotation-equivariant CNN architecture which shares weights over filter orientations to improve generalization and to reduce sample complexity.
A key property of our network is that its filters are learned such that they are steerable.
This approach avoids interpolation artifacts which can be severe at the small length scale of typical filter kernels.
We accomplish the steerability of the learned filters by representing them as linear combinations of a fixed system of atomic steerable filters.

In all intermediate layers of the network, we utilize group convolutions to ensure an equivariant mapping of feature maps.
Group-convolutional networks were proposed by \citet{cohen2016group} who considered four filter orientations.
An advantage of our construction is that we can achieve an arbitrary angular resolution w.r.t.\@ the sampled filter orientations.
Indeed, our experiments show that results improve significantly when using more than four orientations.

\begin{figure*}[t]
	\centering
	\includegraphics[trim=0mm 2mm 0mm 0mm, width=1.\textwidth]{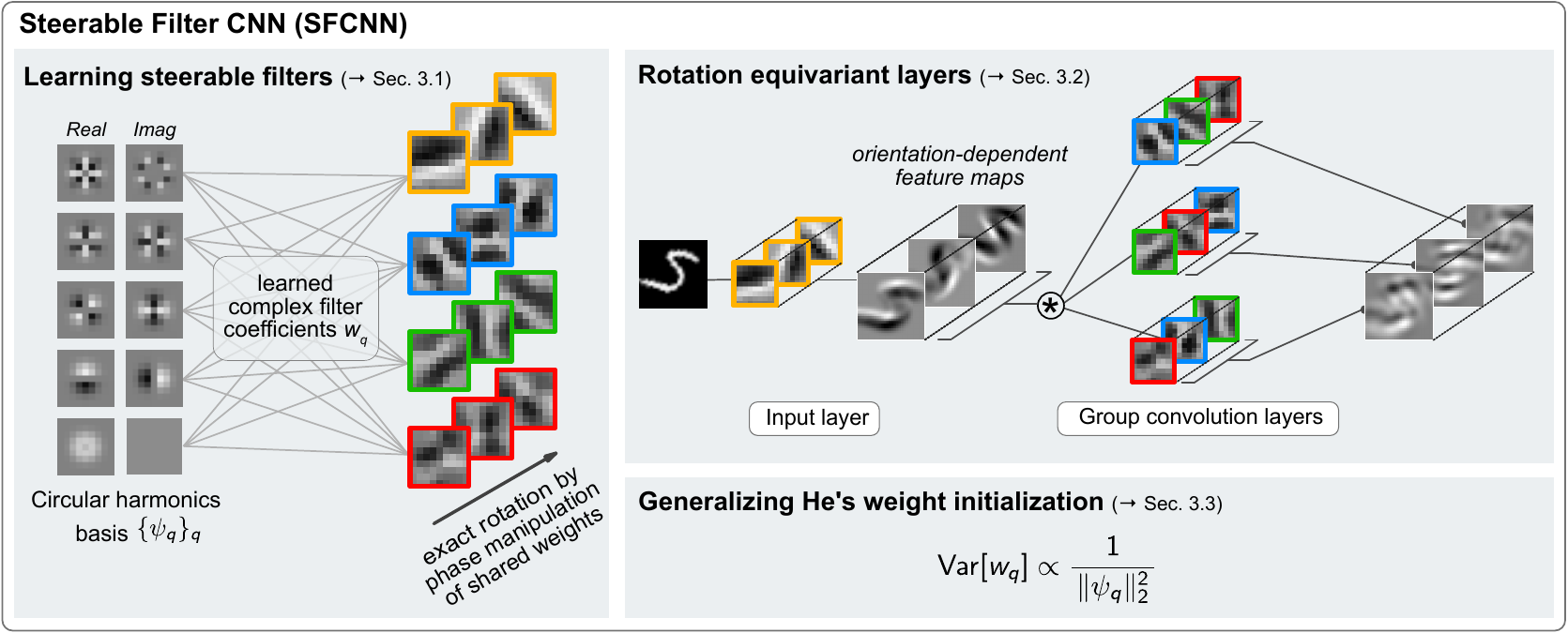}
	\caption{
		Key concepts of the proposed Steerable Filter CNN:
		The filters are parameterized in a steerable function space with shared weights over filter orientations.
		Exact filter rotations are achieved by a phase manipulation of the expansion coefficients $w_q.$
		All layers are designed to be jointly translation and rotation equivariant.
		The weights $w_q$ serve as expansion coefficients of a fixed filter basis $\{\psi_q\}_{q}$ rather than pixel values.
		Therefore, we adapt He's weight initialization scheme to this more general case which implies to normalize the basis filter energies.
	}
	\vspace*{-2.5ex}
	\label{fig:overview}
\end{figure*}

An important practical aspect of CNNs is a proper weight initialization.
Since the weights to be learned serve as expansion coefficients for the steerable function space, common weight initialization schemes need to be adapted.
Here, we generalize the results found by \citet{Glorot10understandingthe} and \citet{DBLP:journals/corr/HeZR015} to networks which learn filters as a composition of (not necessarily steerable) atomic filters.

Our network achieves state-of-the-art results on two important rotation-equivariant/invariant recognition tasks: 
(\emph{i})~The proposed approach is the first to obtain an accuracy higher than 99\% on the rotated MNIST dataset, which is the standard benchmark for rotation-invariant classification.
(\emph{ii})~A~processing pipeline based on the proposed \mbox{SFCNN} layers ranks among the top three entries in the ISBI 2012 electron microscopy segmentation challenge \cite{arganda2015crowdsourcing}.

Figure~\ref{fig:overview} gives an overview over the key concepts utilized in Steerable Filter \mbox{CNNs}.

\section{Equivariance properties of CNNs}

Equivariance is the property of a function to commute with the actions of a symmetry group acting on its domain and codomain.
Formally, given a transformation group $G,$ a function $f:X\to Y$ is said to be equivariant if
\begin{equation*}
	f\left(\varphi_g^X (x)\right)\ =\ \varphi_g^Y \left(f\left(x\right)\right) \quad \forall g\in G,\ x\in X,
\end{equation*}
where $\varphi_g^{(\cdot)}$ denotes a \emph{group action} in the corresponding space.
A special case of equivariance is invariance for which $\varphi_g^Y=\id$.

In many machine learning tasks a set of transformations is known a-priori under which the prediction should transform in an equivariant way.
Including such knowledge directly into the model can greatly facilitate learning by freeing up model capacity for other factors of variation.
As an example consider a segmentation problem where the goal is to learn a mapping from an image space $\mathcal{I}$ to label images in $\mathcal{L}$, which we formalize by a ground truth segmentation map $\mathcal{S}:\mathcal{I}\to\mathcal{L}.$
The learning process involves fitting a model $\mathcal{M}:\mathcal{I}\to\mathcal{L}$ to approximate the ground truth.
For segmentation tasks, however, translations of the input image $I\in\mathcal{I}$ should typically lead to a translated segmentation map.
Specifically, one has
\begin{equation}\label{eq:translation_equivariance}
	\mathcal{S}\left(\mathcal{T}_d I\right) = \mathcal{T}_d\mathcal{S}(I) \quad \forall d\in \mathbb{R}^2,\ \ I\in \mathcal{I},
\end{equation}
where $\mathcal{T}_d$ is an action of the translation group $T=(\mathbb{R}^2,+)$ which shifts the image or segmentation by $d\in\mathbb{R}^2.$
The group action partitions the image space in equivalence classes $T.I=\{\mathcal{T}_d I \,|\, d\in\mathbb{R}^2\}$ which are known as \emph{group orbits} and comprise all images that are related by the action.
Note that the translation equivariance \eqref{eq:translation_equivariance} of the ground truth segmentation function implies a mapping of whole orbits in $\mathcal{I}$ to orbits in $\mathcal{L}.$
It is therefore possible to reformulate the ground truth as $\widetilde{\mathcal{S}}:\mathcal{I}/T\to\mathcal{L}/T$,
where $(\cdot)/T$ denotes the \emph{quotient space} resulting from collapsing equivalent images in an orbit to a single element.
Instead of fitting an unrestricted model $\mathcal{M}$ to $\mathcal{S}$ it is advantageous to incorporate the transformation behavior into the model by construction.
The crucial consequence is that this reduces the hypothesis space to models $\widetilde{\mathcal{M}}:\mathcal{I}/T\to\mathcal{L}/T.$

CNN layers, which transform feature maps $\zeta$ by convolving them with filters $\Psi,$ are by construction equivariant under translations, that is, $\left(\mathcal{T}_d \zeta\right) \ast \Psi = \mathcal{T}_d\left(\zeta \ast \Psi\right).$
Therefore, their hypothesis space is restricted to $\widetilde{M}.$
\footnote{In practice one often uses strided pooling layers which make the prediction more robust to local deformations but reduce the equivariance to a subgroup determined by the stride.}
As consequence, patterns learned at one specific location evoke the same response at each other location which leads to reduced sample complexity and enhanced generalization.

Besides translations, there are often further transformations like rotations, mirroring or dilations under which the model should be equivariant.
Enforcing equivariance under an extended transformation group $G$ leads to an enhanced generalization over larger orbits $G.I$ and reduces the hypothesis space further to $\widetilde{\mathcal{M}}:\mathcal{I}/G\to\mathcal{L}/G.$

\section{Steerable Filter CNNs}

Here, we develop Steerable Filter CNNs (\mbox{SFCNNs}) which achieve equivariance under joint translations and discrete rotations.
The key concept leading to translation equivariance of CNNs is \emph{translational weight sharing}.
We extend the transformation group under which our networks' layers are equivariant by additionally \emph{sharing weights over filter orientations}.
This implies to perform convolutions with several rotated versions of each filter.
The rotational weight sharing leads to an improved sample complexity and to an enhanced generalization over orbits consisting of images connected by translations and discrete rotations.

In the following sections we introduce our parametrization of steerable filters, propose the network design in terms of these filters and derive a weight initialization scheme adapted to the filter parametrization.
For the formal derivations we assume the images, feature maps and filters to be defined on the continuous domain $\R^2$.
The effects resulting from a discretized implementation are investigated in the experimental section.

\subsection{Parametrization of the steerable filters}\label{sec:FilterConstruction}

At the heart of convolutional neural networks lies the concept of learning filter kernels.
Our construction demands for filters whose responses can be computed accurately and economically for several filter orientations.
Simultaneously the filters should not be restricted in their expressive power, i.e. in the patterns to be learned.
All of these requirements are met by learning linear combinations of a system of steerable filters.
Here we describe a suitable construction of steerable filters for learning in CNNs.

A filter $\Psi$ is rotationally steerable in the sense of \citet{hel1998canonical}, when its rotation by an arbitrary angle $\theta$ can be expressed in a function space spanned by a fixed set of atomic basis functions $\{\psi_q\}_{q=1}^Q.$
This definition includes the classical formulation of steerability by \citet{freeman1991design} as a specific choice of basis.
Formally, a steerable filter $\Psi:\R^2\to\R$ satisfies
\begin{equation}\label{eq:steerabilityGeneralized}
	\rho_\theta \Psi(x) = \sum\nolimits_{q=1}^{Q} \kappa_q(\theta) \psi_q(x),
\end{equation}
for all angles $\theta\in(-\pi,\pi]$ and for angular expansion coefficient functions $\kappa_q$.
Here $\rho_\theta$ denotes both the rotation operator defined by $\rho_\theta \Psi(x) = \Psi(\rho_{-\theta}x)$ when acting on a function as well as a counterclockwise rotation by the angle~$\theta$ when acting on a coordinate vector.
As pointed out by \citet{freeman1991design}, the rotation by steerability is analytic and exact even for signals sampled on a grid.
In contrast to rotations by interpolation the approach does not suffer from interpolation artifacts.
An important practical consequence of steerability is that the response of each orientation can be synthesized from the atomic responses $f\ast\psi_q$; that is,
$
	\left(f \ast \rho_\theta \Psi\right)(x) = \sum\nolimits_{q=1}^{Q} \kappa_q(\theta) \left(f \ast \psi_q\right)(x).
$

\begin{figure}
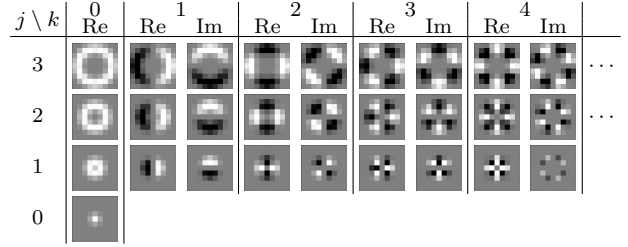

	\centering
	\def\figfolder{images/atoms/}
	\def\figheight{0.072\columnwidth}
	\renewcommand{\tabcolsep}{2.2pt}
	\newlength{\twidth}
	\newlength{\myskip}
	\newlength{\myskipS}
	\newlength{\vertSkip}
	\setlength\twidth{\figheight}
	\setlength\myskip{10pt}
	\setlength\myskipS{3pt}
	\setlength\vertSkip{1.5ex}
	\renewcommand{\arraystretch}{.4}
	\newcommand\Tstrut{\rule{0pt}{3.6ex}}		
	\newcommand\Bstrut{\rule[-0.9ex]{0pt}{0pt}}	
	\centering
	\footnotesize
	\begin{tabular}{ c | c | *{4}{ c c | } c}
		 \multirow{2}{*}{$j \setminus k$} & \multicolumn{1}{c|}{$0$} & \multicolumn{2}{c|}{$1$} & \multicolumn{2}{c|}{$2$} & \multicolumn{2}{c|}{$3$} & \multicolumn{2}{c|}{$4$} \\
		 & $\real$ & $\real$ & $\imag$ & $\real$ & $\imag$ & $\real$ & $\imag$ & $\real$ & $\imag$ & \\
		 \hline
		 \Tstrut
		 $3$      &
		 \includegraphicstotab[height=\figheight]{\figfolder atom_r3_k0.png} &
		 \includegraphicstotab[height=\figheight]{\figfolder atom_r3_k1_re.png} &
		 \includegraphicstotab[height=\figheight]{\figfolder atom_r3_k1_im.png} &      
		 \includegraphicstotab[height=\figheight]{\figfolder atom_r3_k2_re.png} &
		 \includegraphicstotab[height=\figheight]{\figfolder atom_r3_k2_im.png} &
		 \includegraphicstotab[height=\figheight]{\figfolder atom_r3_k3_re.png} &
		 \includegraphicstotab[height=\figheight]{\figfolder atom_r3_k3_im.png} &
		 \includegraphicstotab[height=\figheight]{\figfolder atom_r3_k4_re.png} &
		 \includegraphicstotab[height=\figheight]{\figfolder atom_r3_k4_im.png} & $\mathbf{\cdots}$
		 \\[\vertSkip]
		 $2$      &
		 \includegraphicstotab[height=\figheight]{\figfolder atom_r2_k0.png} &
		 \includegraphicstotab[height=\figheight]{\figfolder atom_r2_k1_re.png} &
		 \includegraphicstotab[height=\figheight]{\figfolder atom_r2_k1_im.png} &      
		 \includegraphicstotab[height=\figheight]{\figfolder atom_r2_k2_re.png} &
		 \includegraphicstotab[height=\figheight]{\figfolder atom_r2_k2_im.png} &
		 \includegraphicstotab[height=\figheight]{\figfolder atom_r2_k3_re.png} &
		 \includegraphicstotab[height=\figheight]{\figfolder atom_r2_k3_im.png} &
		 \includegraphicstotab[height=\figheight]{\figfolder atom_r2_k4_re.png} &
		 \includegraphicstotab[height=\figheight]{\figfolder atom_r2_k4_im.png} & $\mathbf{\cdots}$
		 \\[\vertSkip]
		 $1$      &
		 \includegraphicstotab[height=\figheight]{\figfolder atom_r1_k0.png} &
		 \includegraphicstotab[height=\figheight]{\figfolder atom_r1_k1_re.png} &
		 \includegraphicstotab[height=\figheight]{\figfolder atom_r1_k1_im.png} &      
		 \includegraphicstotab[height=\figheight]{\figfolder atom_r1_k2_re.png} &
		 \includegraphicstotab[height=\figheight]{\figfolder atom_r1_k2_im.png} &
		 \includegraphicstotab[height=\figheight]{\figfolder atom_r1_k3_re.png} &
		 \includegraphicstotab[height=\figheight]{\figfolder atom_r1_k3_im.png} &
		 \includegraphicstotab[height=\figheight]{\figfolder atom_r1_k4_re.png} &
		 \includegraphicstotab[height=\figheight]{\figfolder atom_r1_k4_im.png} & 
		 \\[\vertSkip]
		 $0$      &
		 \includegraphicstotab[height=\figheight]{\figfolder atom_r0_k0.png} & 
	\end{tabular}
	\vspace{-1.5ex}
	\caption{
		Illustration of the circular harmonics $\psi_{jk}(r, \phi) = \tau_j(r) \, e^{ik\phi}$ sampled on a  $9 \times 9$ grid.
		Each row shows a different radial part~$j$, the angular frequencies are arranged in the columns.
		For larger scales there are higher frequency filters not shown here.
	}
	\label{fig:steerableFilterAtoms}
	\vspace{-3.5ex}
\end{figure}

A basis of a steerable function space which is particularly easy to handle is given by circular harmonics; see e.g. \cite{Hsu:82, Rosen:88}.
They are defined by a sinusoidal angular part multiplied with a radial function $\tau:\R^+\to\R$, i.e.
\vspace{-.4ex}
\begin{equation}\label{eq:polarSeparableConstruction}
	\psi_{k}(r, \phi)\ =\ \tau(r) \, e^{\ci k\phi},
\end{equation}
where $(r, \phi)$ denote polar coordinates of $x = (x_1, x_2)$ and $k\in\Z$ is the angular frequency.
By construction, $\psi_{k}$ can be rotated by multiplication with a complex exponential,
\vspace{-.4ex}
\begin{equation}
	\rho_\theta \psi_{k}(x)\ =\ e^{-\ci k\theta} \psi_{k}(x).
\end{equation}
In our network, we utilize a system of circular harmonics  $\psi_{jk}$ with  $j=1,\dots,J,$ and $k=0,\dots,K_j$  where the additional index $j$ controls the radial part of $\psi_{jk} = \tau_j(r) \, e^{\ci k\phi}$.
Figure~\ref{fig:steerableFilterAtoms} shows the real and imaginary parts of the atoms used in the experiments where we chose Gaussian radial parts $\tau_j(r)=\exp({-(r-\mu_j)^2}/{2\sigma^2})$ with $\mu_j=j.$
The maximum angular frequencies $K_j$ are limited to the point where aliasing effects occur.
We found this system to be convenient for learning as the filters are approximately orthogonal and radially localized.

The learned filters are then defined as linear combinations of the elementary filters, that is,
\vspace{-.4ex}
\begin{equation}\label{eq:combinedFilter}
	\tilde{\Psi}(x)\ =\ \sum\nolimits_{j=1}^{J} \sum\nolimits_{k=0}^{K_j} w_{jk} \psi_{jk}(x),
\end{equation}
with weights $w_{jk} \in \C.$ The complex phase of the weights allows rotating the atomic filters \emph{with respect to each other}.
Such a composed filter can subsequently be steered \emph{as a whole} by phase manipulation of the atoms via
\vspace{-.4ex}
\begin{equation}
	\rho_\theta \tilde{\Psi}(x) = \sum\nolimits_{j=1}^{J} \sum\nolimits_{k=0}^{K_j} w_{jk} e^{-\ci k\theta} \psi_{jk}(x).
\end{equation}
We select a single orientation
by taking their real part
\vspace{-.4ex}
\begin{equation}\label{eq:learnedFilterReal}
	\Psi(x)\ =\ \real\tilde{\Psi}(x)
\end{equation}
and let $\rho_\theta\Psi = \real\rho_\theta\tilde{\Psi}$.

\subsection{Equivariant network architecture}

\noindent
The basic building blocks of the proposed SFCNN are three equivariant layer types which we introduce in this section.

\myparagraph{Input layer:}
The first layer $l=1$ of our network ingests an image with $C$ channels $I_c:\mathbb{R}^2\to\mathbb{R},\ c=1,\ldots C$ and convolves these with $\hat{C}$ rotated filters $\rho_\theta \Psi_{\hat{c}c}^{(1)},$ where $\Psi_{\hat{c}c}^{(1)}:\R^2\to\R,\ \hat{c}=1,\ldots,\hat{C},$ are filter channels of the form~\eqref{eq:learnedFilterReal}.
This results in pre-nonlinearity features
 \begin{align}\label{eq:firstLayer}
	&\ y_{\hat{c}}^{(1)}(x,\theta)\ =\ \sum\nolimits_{c=1}^{C} \left(I_c \ast \rho_\theta\Psi_{\hat{c}c}^{(1)}\right)(x) \\
	=&\ \sum\nolimits_{c=1}^{C}  \left(I_c \ast \real \sum\nolimits_{j=1}^{J} \sum\nolimits_{k=0}^{K_j}w_{\hat{c}cjk} e^{-\ci k\theta} \psi_{jk}\right)(x)\notag\\
	=&\ \real \sum\nolimits_{c=1}^{C}\sum\nolimits_{j=1}^J\sum\nolimits_{k=0}^{K_j} w_{\hat{c}cjk} e^{-\ci k\theta} \left(I_c * \psi_{jk}\right)(x)\notag,
\end{align}
where the filters are rotated by in total $\Lambda$ equidistant orientations $\theta \in \Theta = \{ 0, \ldots, 2\pi\frac{\Lambda-1}{\Lambda} \}$.
In this setting the rotational weight sharing is reflected by the phase manipulation of the weights $w_{\hat{c}cjk}$ which themselves are independent of the angle $\theta$.
A higher resolution in orientations can be achieved by simply expanding the tensor containing the phase-factors.

As usual, after the convolution step a bias $\beta^{(1)}_{\hat{c}}$ is added and a nonlinearity $\sigma$ is applied, so that we end up with the first layer's feature map given by
\[
	\zeta^{(1)}_{\hat{c}}(x, \theta)\ =\ \sigma\left( y_{\hat{c}}^{(1)}(x,\theta) + \beta^{(1)}_{\hat{c}} \right).
\]
Note that the resulting representation $\zeta^{(1)}_{\hat{c}}$ depends on a spatial location $x$ and an orientation angle $\theta,$ i.e. on the transformation group applied to the filters.

\myparagraph{Group-convolutional layers:}
To process the resulting feature maps further we utilize group convolutions which naturally generalize spatial convolutions from translations to more general transformation groups.
Given a feature map $\zeta:G\to\mathbb{R}$ and a filter $\Psi:G\to\mathbb{R}$ living on a group $G$, their group convolution is defined by
$
	(\zeta \circledast \Psi)(g) = \int_{G} \zeta(h)\Psi(h^{-1}g) \, d\lambda(h),
$
where we use the symbol $\circledast$ to distinguish group convolutions from the spatial convolution operator $\ast$, and $\lambda$ denotes a Haar measure.
The resulting feature map is again a function on the group.
In analogy to spatial convolutions, group convolutions are equivariant under the group operation, i.e.
$
	\left(\varphi_h(\zeta) \circledast \Psi\right) (g) = \varphi_h \left(\zeta \circledast \Psi\right) (g), \quad \forall h, g \in G,
$
where  $\varphi_h$ is given by $\varphi_h \zeta(g) = \zeta(h^{-1}g).$
For a deeper discussion of group convolutions in neural networks we refer to \cite{cohen2016group}.

The feature maps calculated by the input layer are functions on the semidirect product group $\R^2\rtimes\Theta\leq\text{SE}(2).$
Keeping the parameterization by $(x,\theta),$ the group convolutions 
with summation over input channels 
can be explicitly instantiated as
\allowdisplaybreaks
\begin{align}\label{eq:groupConvExpandedO2}
	&\ y_{\hat{c}}^{(l)}(x,\theta)\ =\ \sum\nolimits_{c=1}^{C}\left(\zeta_c^{(l-1)} \circledast \Psi_{\hat{c}c}^{(l)}\right)(x, \theta) \\
	=& \sum\nolimits_{c=1}^{C}\sum\nolimits_{\phi \in \Theta} \int_{\R^2} \zeta_c^{(l-1)}(u, \phi)\Psi_{\hat{c}c}^{(l)}\left((u,\phi)^{-1}(x,\theta)\right) du \notag\\
	=& \sum\nolimits_{c=1}^{C}\sum\nolimits_{\phi \in \Theta} \left(\zeta_c^{(l-1)}(\cdot, \phi) * \rho_{\phi}\Psi_{\hat{c}c}^{(l)}(\cdot, \theta - \phi)\right)(x) \notag\\
	=& \sum\nolimits_{c=1}^{C}\sum\nolimits_{\phi \in \Theta} \left(\zeta_c^{(l-1)}(\cdot, \phi) * \mathcal{R}_{\phi}\Psi_{\hat{c}c}^{(l)}(\cdot, \theta)\right)(x).\notag
\end{align}
Here the multiplication with the inverse group element, $(u,\phi)^{-1}(x,\theta) = (\rho_{-\phi}(x-u), \theta-\phi)$, was evaluated by switching to a representation of the group.
We further introduced the action $\mathcal{R}_{\phi}$ defined by
\[
	\mathcal{R}_{\phi} \Psi(x,\theta) \ :=\ \rho_{\phi} \Psi(x,\theta-\phi)
\]
which transforms functions on the group by rotating them spatially and shifting their orientation components cyclically.
The above equation reveals that the group convolution can be decomposed into a spatial convolution, rotation and linear combination.
In analogy to the first layer we make use of the steerable filters which on the group are defined by
$
	\Psi_{\hat{c}c}^{(l)}(x, \theta) = \real \sum\nolimits_{j=1}^{J} \sum\nolimits_{k=0}^{K_j} w_{\hat{c}cjk\theta} \psi_{jk}(x).
$
Note that the additional orientation dimension is reflected by an additional index of the weight tensor.
Inserting the steerable filters in \eqref{eq:groupConvExpandedO2} we obtain the pre-nonlinearity feature maps of the group-convolutional layers
\begin{align}\label{eq:groupConvFilter}
	&\ y^{(l)}_{\hat{c}}(x,\theta) \\
	=& \sum_{c=1}^{C} \sum_{\phi\in\Theta} \left( \zeta_c^{(l-1)}(\cdot, \phi) \ast \real \sum_{j,k} w_{\hat{c}cjk,\theta-\phi} e^{- \ci k\phi} \psi_{jk}\right)\!\!(x)\notag\\
	=& \real \sum_{c=1}^{C} \sum_{\phi\in\Theta} \sum_{j,k} w_{\hat{c}cjk,\theta-\phi} e^{- \ci k\phi} \left(\zeta_c^{(l-1)}(\cdot, \phi) * \psi_{jk}\right)\!(x).\notag
\end{align}
As before, a bias $\beta^{(l)}_{\hat{c}}$ is added and the activation function $\sigma$ is applied,
$\zeta^{(l)}_{\hat{c}}(x,\theta) = \sigma(y^{(l)}_{\hat{c}}(x,\theta) + \beta^{(l)}_{\hat{c}}).$

By the linearity of the steerability and the convolution, one can implement the layers either by a direct convolution with linearly combined filters, or by linearly combining the responses of the atomic filters.
We implemented both approaches and found that in typical operation regimes the first option is faster since the kernels to be linearly combined have a smaller spatial extent than the atomic responses of the second option.

\begin{figure*}
	\centering
	\includegraphics[width=1\textwidth]{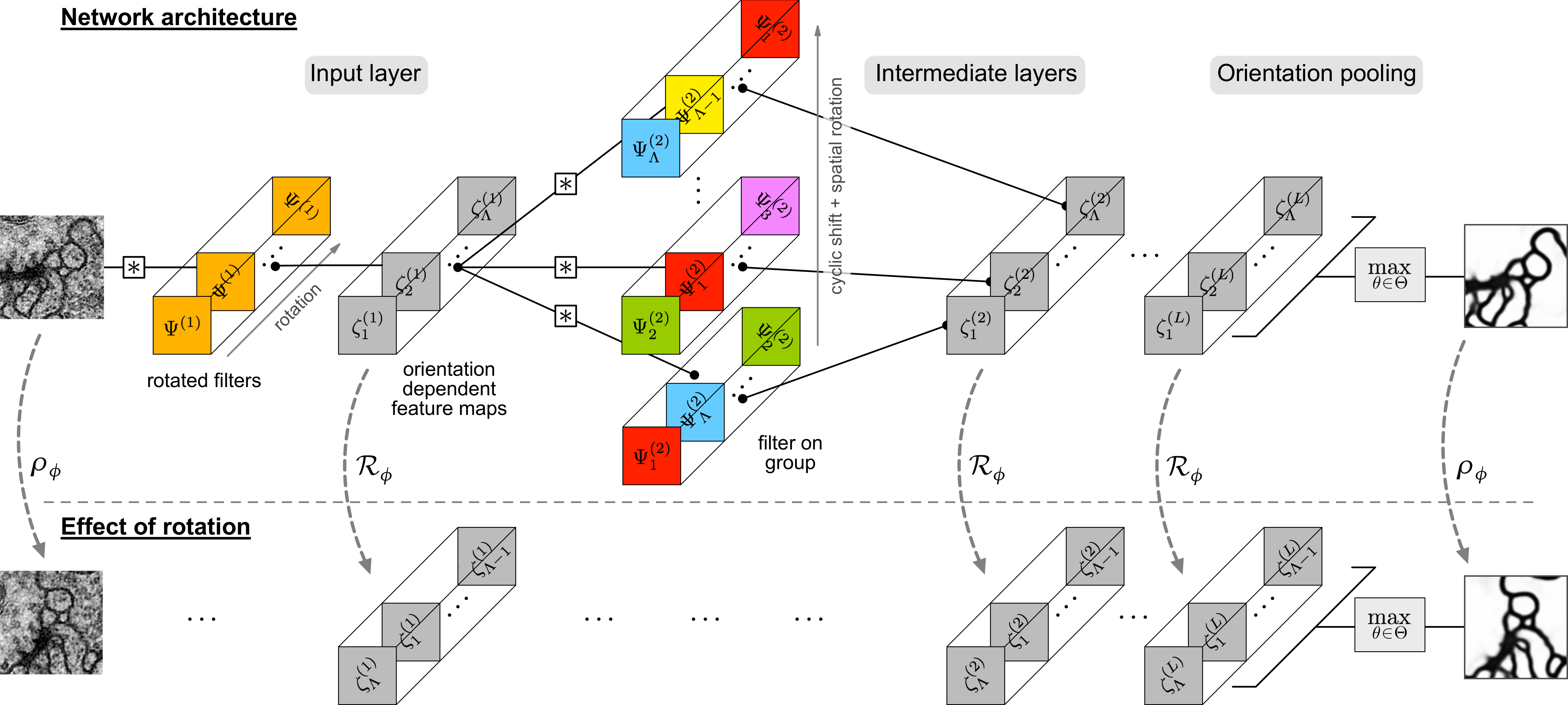}
	\vspace{-3.5ex}
	\caption{
		\emph{Top:} Basic structure of a typical \mbox{SFCNN} for rotation-equivariant segmentation.
		For clarity, we display only a single group-convolutional layer and a single feature channel and omit pooling and normalization layers.
		Rotated Greek letters represent the spatial orientations of the filters and the feature maps.
		Orientation components are abbreviated as subscript, i.e. $\Psi_{\lambda} = \Psi(\cdot, \theta_\lambda).$
		Filters in the same color share their weights as they are connected by rotations.
		The weight sharing of the filters on the group is prescribed by the group convolution \eqref{eq:groupConvExpandedO2}.
		After the last group-convolutional layer we pool over orientations to obtain predictions which are invariant under rotations of local patches in the field of view.
		\emph{Bottom:}~Visualization of the layerwise rotation-equivariance.
		Applying a rotation $\rho_\phi$ to the input image results in a joint spatial rotation operation and cyclic shift over orientation indices $\mathcal{R}_\phi$ of the feature maps $\zeta^{(l)}.$
		This transformation behavior can be understood intuitively when paying attention to the relative orientation of each layer's input and filters.
		}
	\label{fig:network}
	\vspace{-3ex}
\end{figure*}

\myparagraph{Output layer:}
After the last group-convolutional layer we extract the information of interest for the specific task.
For rotation-invariant classification we pool globally over both the orientation dimension and the remaining spatial resolution.
A pooling over orientations is also done for rotation-equivariant segmentation where spatial dimensions remain and the output rotates according to the rotation of the network's input.
If the orientation itself is of interest it could be kept as extra feature.

\myparagraph{Equivariance:}
Each individual layer $L_{(\cdot)}$ of the network is equivariant under joint translations and rotations in the group $\mathbb{R}^2\rtimes\Theta:$
Rotating the input image leads to a transformation
$L_{\text{in}}(\rho_\phi I) = \mathcal{R}_\phi L_{\text{in}}(I)$
of the first layer's feature maps.
The subsequent group-convolutional layers then transform like
$L_{\text{gconv}}(\mathcal{R}_\phi \zeta) = \mathcal{R}_\phi L_{\text{gconv}}(\zeta).$
When using orientation pooling in the output layer the resulting feature maps are rotated:
$L_{\text{out}}(\mathcal{R}_\phi \zeta) = \rho_\phi L_{\text{out}}(\zeta).$
Overall, this implies the equivariance of a whole network,
\[
	\left(L_{\text{out}} \circ L_{\text{gconv}}^d \circ L_{\text{in}}\right)(\rho_\phi I) = \rho_\phi \left(L_{\text{out}} \circ L_\text{gconv}^d \circ L_{\text{in}}(I)\right),
\]
where $d$ is the number of group-convolutional layers.
The layers' equivariance is formally proven in appendix~\ref{apx:Equivariance}.

The top part of Figure~\ref{fig:network} visualizes the building blocks of a typical \mbox{SFCNN} for rotation-equivariant segmentation.
An overview over the transformation behavior of the feature maps under rotation of the input is given in the bottom part.
The spatial rotation and cyclic shift over orientation channels $\mathcal{R}_\phi$ of the feature maps on the group can be understood intuitively when paying attention to the relative orientation of each layer's input and filters.

Compared to a conventional CNN which independently learns filters in $\Lambda$ orientations in a rotation-invariant recognition task, a corresponding \mbox{SFCNN} consumes $\Lambda$ times less parameters to extract the same representation.

SFCNN incur a small computational overhead for building the filter kernels from the circular harmonics basis which we found to be negligible.
The computational cost of SFCNNs is therefore equivalent to that of a conventional CNN when the effective number of channels coincide, i.e. when $I_\text{CNN}=\Lambda I_\text{SFCNN}.$

\subsection{Generalizing He's weight initialization scheme}\label{sec:weightInitialization}

An important practical aspect of training deep networks is an appropriate initialization of their weights.
When the weights' variances are chosen too high or low, the signals propagating through the network are amplified or suppressed exponentially with depth.
\citet{Glorot10understandingthe} and \citet{DBLP:journals/corr/HeZR015} investigated this issue and came up with initialization schemes which are accepted as a standard for random weight initialization.
In contrast to \cite{Glorot10understandingthe} and \cite{DBLP:journals/corr/HeZR015} our filters are not parameterized in a pixel basis but as a linear combination of a system of atomic filters with weights serving as expansion coefficients.
To be specific, we consider filters $\Psi_{\hat{c}cx}=\sum_{q=1}^Q w_{\hat{c}cq} \psi_{qx}$ which are built from $Q$, not necessarily steerable, real valued atomic filters which map $C$ input channels to $\hat{C}$ output channels.
This assumption is more general than that of the aforementioned works since they only consider the pixel basis $\psi_{qx}^\text{Dirac}=\delta_{q,x}$, i.e. atomic filters which are zero everywhere but at one pixel.

Most of the further assumptions are identical to those in~\cite{DBLP:journals/corr/HeZR015}:
We assume the activations and gradients to be i.i.d. and to be independent from the weights.
Further, the weights themselves are initialized to be mutually independent and have zero mean.
An important difference is that we do \emph{not} restrict the weights to be identically distributed because of the inherent asymmetry of the different atomic filters.
All biases are initialized to be zero and the nonlinearities are chosen to be ReLUs.
These assumptions lead to the initialization conditions
\[
	\Var\left[w_q\right]=\frac{2}{CQ\norm{\psi_q}_2^2} \quad \text{or} \quad \Var\left[w_q\right]=\frac{2}{\hat{C}Q\norm{\psi_q}_2^2}
\]
for the forward or backward pass, respectively.
A detailed derivation is given in appendix~\ref{apx:HeWeightDeriv}.

As discussed in \cite{DBLP:journals/corr/HeZR015}, the difference between both initializations cancels out for intermediate layers.
Note that our results include those of \citet{DBLP:journals/corr/HeZR015}, that is, $\Var\left[w_q\right]=\frac{2}{n_\text{in}}$ or $ \Var\left[w_q\right]=\frac{2}{n_\text{out}},$ for $\psi_{qx}^\text{Dirac}=\delta_{q,x}$ with $\norm{\psi_{qx}^\text{Dirac}}_2^2=1.$
We further want to point out that the learned filters are combined of products $w_q\psi_q$ which implies that the factors $\norm{\psi_q}_2^2$ counterbalance different energies of the basis filters.
A convenient way to initialize the network is hence to normalize all filters to unit norm and subsequently initialize the weights uniformly by $\Var\left[w_q\right]=\frac{2}{CQ}$ or $\Var\left[w_q\right]=\frac{2}{\hat{C}Q}.$
In our group-convolutional layers the filters additionally comprise orientation channels.
From the perspective of weight initialization these have the same effect as conventional channels, therefore we propose to normalize their weights variance with an additional factor of $\Lambda.$
We emphasize that using normalization layers like batch normalization does not obviate the need for a proper weight initialization.
This is because such layers scale activations as a whole while our initialization conditions indicates that the relative scale of the summands contributing to each activation needs to be adapted.
Further details, in particular on initializing weights of complex-valued filters, are given in the appendix.

\section{Prior and related work}

A priori knowledge about transformation-invariance of images can be exploited in manifold ways.
A commonly utilized technique is data augmentation, see e.g. \cite{krizhevsky2012imagenet}.
The basic idea is to enrich the training set by transformed samples.
Augmenting datasets allows to train larger models and is easily applicable without modifying the network architectures.
When the augmenting transformations form a group $G$ the additional images $I\subseteq\mathcal{I}$ lie on the orbit $G.I$.
In contrast to equivariant models the hypothesis space is not restricted to the quotient space $\mathcal{I}/G$ under the utilized symmetry group but the equivariance needs to be learned explicitly by the network.
This demands for a high learning capacity which makes the network prone to overfitting.

Recent work focuses on incorporating equivariance to various transformations directly into the network's architecture.
Invariance to specific transformations can be achieved by applying them to the input and subsequently pooling their responses \cite{ICML2012Sohn_659, DBLP:journals/corr/KanazawaSJ14, zhang2015discriminative}.
In \cite{NIPS2014_5424} the regions in symmetry space to pool over are learned to become invariant only to nuisance deformations.
Another approach is to resample the input and apply standard convolutions.
\citet{pmlr-v70-henriques17a} achieve equivariance w.r.t.~Abelian symmetry groups by fixing a sampling grid according to the symmetry while in \cite{DBLP:journals/corr/JaderbergSZK15,lin2016inverse} the network itself estimates the grid.
In \cite{jacobsen2017dynamic} transformations are dealt with by convolving with filters which are steered by a subnetwork.

In particular, there has recently been a considerable interest in rotation-equivariant \mbox{CNNs}.
The work \cite{dieleman2016exploiting} introduces four operations which are easily included into existing networks and enrich both the batch- and feature dimension with transformed versions of their content. 
In \cite{cohen2016group}, the feature maps resulting from transformed filters are treated as functions of the corresponding symmetry-group which allows to use group-convolutional layers.
As their computational cost is coupled to the size of the group, \citet{cohen2016steerable} propose to alternatively use steerable representations as composition of elementary feature types.
Besides translations and rotations, the aforementioned works also incorporate reflections, i.e. they operate on the dihedral group.
Their current limitation is the restriction to rotations by the angle $\frac{\pi}{2},$ thus to four orientations.
In \cite{laptev2016ti}, several rotated versions of the same image are sent through a conventional CNN. The resulting features are subsequently pooled over the orientation dimension.
The approach can be easily extended to other transformations.
On the downside, the equivariance is only w.r.t.~global transformations.
\citet{marcos2016learning} perform convolutions with rotated versions of a each filter in a shallow network followed by a global pooling over orientations.
These ideas were extended to networks which additionally propagate the orientation of the maximum response \cite{marcos2016rotation}.
In both approaches the filter rotation is based on bicubic interpolation, allowing for fine resolutions with respect to the orientation but causing interpolation artifacts.
\citet{worrall2016harmonic} achieve continuous resolution in orientations by working with complex valued steerable filters and feature maps.
However, this requires the angular frequencies of the feature maps to be kept disentangled. 
Rotation-equivariant feature extraction can also be achieved by using group-convolutional scattering transforms \cite{sifre2013rotation}.
A fundamental difference to our work is that the filter banks are fixed rather than learned.

\begin{figure*}[!ht]
	\centering
	\includegraphics[width=1.\textwidth]{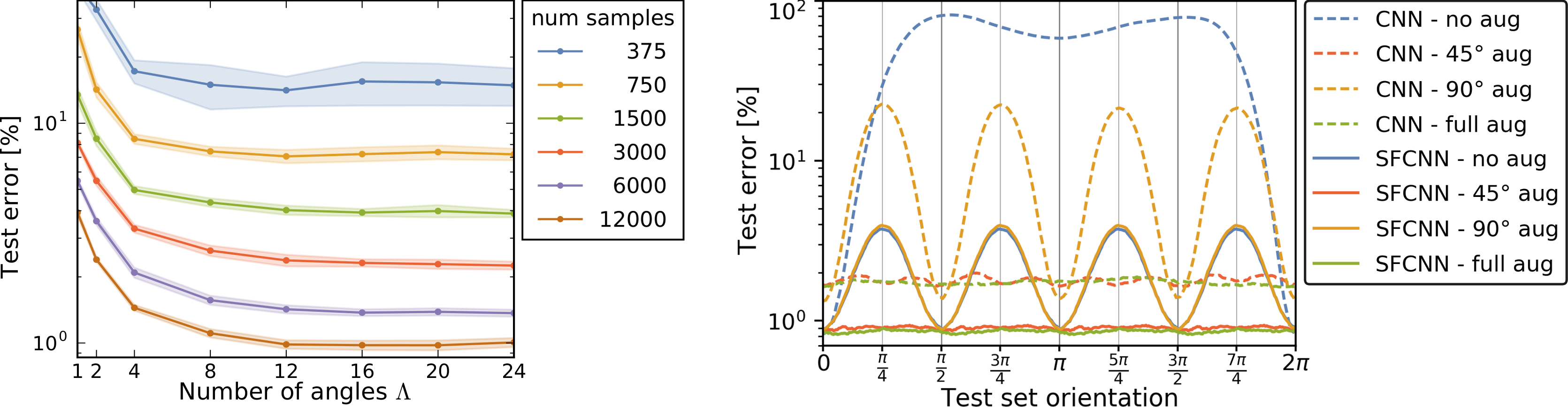}
	\vspace{-3.5ex}
	\caption{
		\emph{Left}:
		Test error versus number of sampled filter orientations for different training subsets from mnist-rot.
		Shaded regions highlight the standard deviations over several runs.
		The accuracy improves significantly with increasing angular resolution until it saturates at around $12$ to $16$ orientations.
		\emph{Right}:
		Rotational generalization capabilities of a conventional CNN and a \mbox{SFCNN} with $\Lambda=16$ using different data augmentation strategies.
		In this experiment the training set consists of unrotated MNIST digits while the test set for each angle contains the remaining digits, rotated to the corresponding angle.
		}
	\label{fig:mnist_experiments_combined}
	\vspace{-2ex}
\end{figure*}

\section{Experimental results}

We evaluate the proposed \mbox{SFCNNs} on two datasets exhibiting rotational symmetries.
On the rotated MNIST dataset we first investigate specific network properties like the accuracy's dependence on the number of sampled orientations and the generalization of learned patterns over orientations.
With the insights gained in these experiments we benchmark the model and the proposed initialization scheme.
To evaluate the segmentation capabilities of \mbox{SFCNNs} on real world data we run a further experiment on the ISBI 2012 EM segmentation challenge.

\subsection{Rotated MNIST}

In our first experiments we investigate the equivariance properties of the proposed network architecture on the
\href{http://www.iro.umontreal.ca/~lisa/twiki/bin/view.cgi/Public/MnistVariations}{rotated MNIST dataset}
(mnist-rot) which is the standard benchmark for rotation-equivariant models.
The dataset contains the handwritten digits of the classical MNIST dataset, rotated to random orientations in $[0,2\pi)$.
It is split in $12000$ training and $50000$ test images; model selection is done by training on $10000$ images and validating on the $2000$ remaining samples in the training set.

For our initial experiments we utilize the classification-\mbox{SFCNN} given in Table~\ref{tab:architectureCNN1} in the appendix as baseline.
It consists of one steerable input layer which maps the input images to the group, five following group-convolutional layers and three fully connected layers.
After every two steerable filter layers we perform a spatial $2\times2$ max-pooling.
The orientation dimension and the remaining spatial dimensions are pooled out globally after the last convolutional layer.
Details on the further training setup are given in appendix~\ref{apx:Experiments}.

\myparagraph{Sampled orientations:}
The number of sampled orientations $\Lambda$ is a parameter specific to our network, so we first explore its influence on the test accuracy.
We are further interested in the network's sample complexity, i.e.~the dependence on the size of the training set.
The accuracies resulting when varying these parameters are reported in Figure~\ref{fig:mnist_experiments_combined} (left).
As expected, the test error and its standard deviation decrease with the size of the training data set.
We observe that the accuracy improves significantly when increasing the number of orientations until it saturates at around $12$ to $16$ angles.
Up to this point, the gain of adding more sampled orientations is considerable.
For example, in almost all cases, increasing the angular resolution from $2$ to $4$ sampled orientations provides a higher gain in accuracy than sticking with $2$ orientations and doubling the number of training samples.
We want to emphasize that the possibility of \mbox{SFCNNs} to go beyond the four sampled orientations of \citep{dieleman2016exploiting, cohen2016group, cohen2016steerable} leads to a significant gain in accuracy.
Note that the case $\Lambda=1$ corresponds, up to the different filter parameterization, to conventional CNNs.

\begin{figure*}[!ht]
	\includegraphics[width=.212\textwidth]{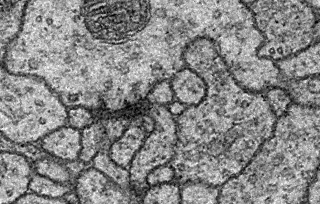}
	\hspace{.4ex}
	\includegraphics[width=.212\textwidth]{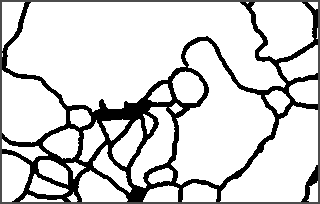}
	\hspace{.4ex}
	\includegraphics[width=.212\textwidth]{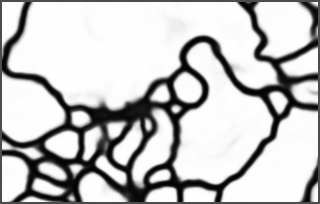}
	\hspace{2.ex}
	\begin{minipage}{0.30\textwidth}
	\vspace{-13.75ex}
	\footnotesize
	\renewcommand{\arraystretch}{.925}
	\begin{tabular}{l@{\hskip 2.5ex}l@{\hskip 2.5ex}l@{\hskip .4ex}}
		\midrule \vspace{-1pt}
		Method							& $V^\text{Rand}$    & $V^\text{Info}$ \\[-1pt]
		\midrule
		IAL MC/LMC						& $\mathbf{0.98792}$ & $\mathbf{0.99183}$ \\
		CASIA\_MIRA						& $0.98788$ & $0.99072$ \\
		\textbf{Ours}					& $0.98680$ & $0.99144$ \\
		\citet{quan2016fusionnet}		& $0.98365$ & $0.99130$ \\
		\citet{beier2017multicut}		& $0.98224$ & $0.98845$ \\
		\citet{drozdzal2016importance}	& $0.98058$ & $0.98816$ \\[-1pt]
		\midrule
	\end{tabular}
	\end{minipage}
	\vspace{-2ex}
	\caption{
		Experimental results on the ISBI 2012 challenge. The shown patches are cropped from slice $30$ of the training data set which we used for validation.
		\emph{Left:} Raw EM image.
		\emph{Mid-left:} Binary membrane ground truth segmentation.
		\emph{Mid-right:} Probability map predicted by the proposed network.
		\emph{Right:} Top 6 of more than 100 entries of the \href{http://brainiac2.mit.edu/isbi_challenge/leaders-board-new}{leaderboard}, accessed on November $13,$ $2017.$ Higher values mean better accuracy.
	}
	\label{fig:ISBI_samples}
	\vspace{-2.4ex}
\end{figure*}

\myparagraph{Rotational generalization:}
In order to test how well the networks generalize learned patterns over orientations we conduct an experiment where we train them on unrotated digits and record their accuracy over the orientation of rotated digits.
Specifically, we take the the first $12000$ samples of the conventional MNIST dataset to train a \mbox{SFCNN} with $\Lambda=16$ as well as a conventional CNN of comparable size using either no augmentation, augmentation by rotations which are multiples of either $\frac{\pi}{4}$ or $\frac{\pi}{2}$ or augmentation by rotations which are densely sampled from $[0,2\pi)$.
As test set we take the remaining $58000$ samples and record the test errors' dependence on the orientation of this dataset.
To obtain a fair comparison between the networks we experiment with conventional CNNs with the same number of parameters or the same number of channels like the \mbox{SFCNN}.
Since both show the same behavior we only report the accuracies of the network with the same number of channels which performs slightly better.
The results are plotted in Figure~\ref{fig:mnist_experiments_combined} (right).
One can see that, lacking rotational equivariance, the conventional CNN does not generalize well over orientations.
When using rotational augmentation the error reduces considerably on average, it grows, however, for small angles in a neighborhood of zero.
This is the case because the network needs to learn to detect the augmented samples additionally which demands an increased learning capacity.
The \mbox{SFCNN} on the other hand generalizes quite well over orientations even without augmentation.
In continuous space we would expect the test error curve to be $\frac{2\pi}{\Lambda}$-periodic because of the rotational equivariance.
The deviations from this behavior can be attributed to the sampling effects of using digitized images.
As to be expected for $\Lambda=16$ orientations, the accuracy is not influenced by augmentation with $\frac{\pi}{2}$-rotations since the additional samples lie on the group orbit on which the network is invariant.
In contrast to conventional CNNs, \mbox{SFCNNs} do not show an increased error for small angles in a neighborhood of zero when using augmentation.
This indicates that the cost of learning rotated versions of each digit is negligible thanks to the approximate rotation equivariance.
An augmentation by rotations which are multiples of $\frac{\pi}{4}$ or by continuous rotations give very similar results.
Both seem to act as a regularization preventing the filters to overfit on the pixel grid.

We conclude that \mbox{SFCNNs} outperform the rotational generalization of CNNs for all levels of augmentation.

\begin{table}[!t]
	\centering
	\footnotesize
	\renewcommand{\arraystretch}{.95}
	\begin{tabular}{l@{\hskip 4.4ex}r@{\hskip .3ex}c@{\hskip .3ex}l}
		\toprule
		Method								& \multicolumn{3}{r}{Test Error (\%)} 	\\
		\toprule
		\textbf{Ours}\ \ --\ \ CoeffInit, train time augmentation & $\mathbf{0.714}$ 	& $\ \mathbf{\pm}\ $ & $\mathbf{0.022}$ \\
		\textbf{Ours}\ \ --\ \ CoeffInit	& $0.880$ 	 & $\ \pm\ $ & $0.029$  \\
		\textbf{Ours}\ \ --\ \ HeInit		& $0.957$ 	 & $\ \pm\ $ & $0.025$ \\
		\citet{marcos2016rotation}\ \ --\ \ test time augmentation & $1.01 \phantom{0}$ &&\\
		\citet{marcos2016rotation} 			& $1.09 \phantom{0}$ &&\\
		\citet{laptev2016ti} 				& $1.2  \phantom{00}$&&\\
		\citet{worrall2016harmonic} 		& $1.69 \phantom{0}$ &&\\
		\citet{cohen2016group} - G-CNN		& $2.28 \phantom{0}$ & $\ \pm\ $ & $0.0004$ \\
		\citet{schmidt2012learning}	 		& $4.0 \phantom{00}$ &&\\
		\citet{ICML2012Sohn_659}			& $4.2 \phantom{00}$ &&\\
		\citet{cohen2016group} - conventional CNN		& $5.03 \phantom{0}$ & $\ \pm\ $ & $0.0020$ \\
		\citet{larochelle2007empirical}		& $10.4\phantom{00}$ & $\ \pm\ $ & $0.27$ \\
		\bottomrule
	\end{tabular}
	\vspace{-1.5ex}
	\caption{
		Test errors on the rotated MNIST dataset.
		We distinguish He initialization (HeInit) from the proposed initialization scheme (CoeffInit).
		}
	\vspace{-3.5ex}
	\label{tab:mnist_rot_results}
\end{table}

\myparagraph{Benchmarking:}
Based on the insights from the above experiments we fix the number of sampled orientations to $\Lambda=16$ and tune the network further to the slightly larger architecture given in Table~\ref{tab:architectureCNN2} in the appendix.
The results are reported in Table~\ref{tab:mnist_rot_results}.
Using the \mbox{SFCNN} with He's weight initialization and no data augmentation, we obtain a test error of $0.957 \%$ which already exceeds the previous state-of-the-art.
The proposed initialization scheme, adapted to filter coefficients, significantly improves the test error to $0.880\%.$
When additionally augmenting the dataset with continuous rotations during training time the error decreases further to $0.714\%.$
To summarize, our approach reduces the best previously published error by a factor of~$29\%.$

\vspace*{-.25ex}
\subsection{ISBI 2012 2D EM segmentation challenge}
\vspace*{-.5ex}

In a second experiment we evaluate the performance of our model on the ISBI 2012 electron microscopy segmentation challenge \cite{arganda2015crowdsourcing}.
The goal of the challenge is to predict the locations of the cell boundaries in the Drosophila ventral nerve cord from EM images which is a key step for investigating the connectome of the brain.
The dataset consists of $30$ train and test slices of size $512\times512$ px with a binary segmentation ground truth provided for the training set.
Figure~\ref{fig:ISBI_samples} shows an exemplary raw EM image with the corresponding ground truth segmentation mask and our network's prediction.
An important property of the dataset is that the images have no preferred orientation which makes it suitable for evaluating rotation-equivariant networks.

We build on an established pipeline introduced in \cite{beier2017multicut} where a crucial step is the boundary prediction via a conventional CNN.
In the present experiment, we replaced their network by a \mbox{SFCNN} with a U-net design \cite{ronneberger2015u}.
The network architecture is visualized in Figure~\ref{fig:unet_architecture} in the appendix.
As loss function we chose a pixel wise binary cross entropy loss.
The dataset was augmented by random elastic deformations, flips and rotations by multiples of $\frac{\pi}{2}$ during train time.
In the experiment on rotational generalization we found that augmenting samples by transformations in a subgroup under which the network is equivariant does not have any effect.
We therefore sampled $\Lambda=17$ orientations which is mutually prime with the $4$ augmented orientations.
This way the augmented images do not fall into a subgroup w.r.t.\@ which the network is invariant.

Segmentation predictions are evaluated by the challenge hosters and ranked w.r.t.\@ the foreground-restricted Rand score $V^\text{Rand}$ and the information score $V^\text{Info}$; for an explanation of these metrics see \cite{arganda2015crowdsourcing}.
The current
\href{http://brainiac2.mit.edu/isbi_challenge/leaders-board-new}{leaderboard}
in Figure~\ref{fig:ISBI_samples} (right) shows that our approach yields top-tier results.
In particular, it improves upon the results of~\cite{beier2017multicut}.

\vspace{-.25ex}
\section{Conclusion}
\vspace{-.5ex}

We have developed a rotation-equivariant CNN whose filters are learned such that they are steerable.
Layerwise equivariance is obtained by using group convolutions.
He's weight initialization scheme is extended to general filter bases which empirically leads to an increased accuracy.
Our network allows sampling an arbitrary number of filter orientations which improves the performance until a saturation is reached.
We confirmed experimentally that \mbox{SFCNNs} generalize learned patterns over orientations and therefore achieve a lower sampling complexity than CNNs in rotation-equivariant recognition tasks.
The proposed \mbox{SFCNNs} achieve state-of-the-art results on rotated MNIST and the ISBI 2012 2D EM segmentation challenge.

\paragraph*{Acknowledgement.}
We would like to thank 
T.~Beier, C.~Pape, N.~Rahaman and I.~Arganda-Carreras for their technical support
and U.~K\"othe and T.~Cohen for valuable discussions.
This work was partially supported by the German Research Foundation (DFG grant STO1126/2-1).

\appendix
\noindent{\LARGE\bfseries Appendix}\\

\section{Equivariance properties}\label{apx:Equivariance}

In this section we prove the equivariance of the individual layers of Steerable Filter CNNs under rotations by the sampled orientations in $\Theta$, assuming signals on a continuous domain $\mathbb{R}^2$.
Translational equivariance follows directly from either the utilization of spatial convolutions or from the independence of the operation on the spatial position.

\subsection{Input layer}

The first layer maps an image $I:\mathbb{R}^2\to\mathbb{R}$ to a feature map $\zeta^{(1)}:\mathbb{R}^2\rtimes\Theta\to\mathbb{R}$ by first convolving it with multiple rotated versions $\rho_\theta\Psi$ of a filter $\Psi:\mathbb{R}^2\to\mathbb{R}$ and subsequently adding a bias $\beta$ and applying a nonlinearity $\sigma$.
Both steps are equivariant under rotations of the image by angles $\alpha\in\Theta$.
This means that $\rho_\alpha I(x)$ is mapped to $\mathcal{R}_\alpha \zeta^{(1)}(x,\theta) = \rho_\alpha \zeta^{(1)}(x,\theta-\alpha)$ where $\mathcal{R}_\alpha$ is the group action on functions on the group.
To see that the first step performs an equivariant mapping, simply insert a rotated image,
\begin{align*}
	\left(\rho_\alpha I \ast \rho_\theta\Psi\right)(x) 
	\ =\ \int\limits_{\R^2} I(\rho_{-\alpha}u)\, \Psi(\rho_{-\theta}(x-u))\, \mathrm{d}u \ , \\
\end{align*}
and substitute $\tilde{u}:=\rho_{-\alpha}u$.
Since the transformation is orthogonal we have $\left|\det\left(\frac{\partial \tilde{u}}{\partial u}\right)\right|=1$ and hence:
\begin{align*}
	\left(\rho_\alpha I \ast \rho_\theta\Psi\right)(x) 
	\ =\ & \int\limits_{\R^2} I(\tilde{u})\, \Psi(\rho_{-(\theta-\alpha)}(\rho_{-\alpha}x-\tilde{u}))\, \mathrm{d}\tilde{u} \\
	=\ & \rho_\alpha \left(I \ast \rho_{\theta-\alpha}\Psi\right)(x) \\
	=\ & \rho_\alpha y^{(1)}(x,\theta-\alpha) \\
	=\ & \mathcal{R}_\alpha y^{(1)}(x,\theta)\ .
\end{align*}
The mutual transformation behavior is visualized in the following commutative diagram:
\begin{center}
	\begin{tikzpicture}
	\node (img) {$I(x)$}; 
	\node[below=1.5475cm of img] (imgrot) {$\rho_\alpha I(x)$};
	\node[right=4cm of img] (featmap) {$y^{(1)}(x,\theta)$}; 
	\node[below=1.5cm of featmap] (featmapact) {$\rho_\alpha y^{(1)}(x,\theta-\alpha)$};

	\path[->] (img) edge node[left]{$\rho_\alpha$} (imgrot);
	\path[->] (img) edge node[above]{$\ast\rho_\theta\Psi$} (featmap);
	\path[->] (featmap) edge node[right]{$\mathcal{R}_\alpha$} (featmapact);
	\path[->] (imgrot) edge node[below]{$\ast\rho_\theta\Psi$} (featmapact);
	\end{tikzpicture}
\end{center}
Adding a bias $\beta$ to each feature map channel and applying a nonlinearity $\sigma$ does not interfere with translational- or rotational equivariance since both operations do neither depend on the spatial position nor orientation channel:
\begin{center}
	\begin{tikzpicture}
	\node (preact) {$y^{(l)}(x,\theta)$}; 
	\node[below=1.5cm of preact] (preactrot) {$\rho_\alpha y^{(l)}(x,\theta-\alpha)$};
	\node[right=4cm of preact] (featmap) {$\zeta^{(l)}(x,\theta)$}; 
	\node[below=1.5cm of featmap] (featmaprot) {$\rho_\alpha \zeta^{(l)}(x,\theta-\alpha)$};

	\path[->] (preact) edge node[left]{$\mathcal{R}_\alpha$} (preactrot);
	\path[->] (preact) edge node[above]{$\sigma(\,\cdot\,+\beta)$} (featmap);
	\path[->] (featmap) edge node[right]{$\mathcal{R}_\alpha$} (featmaprot);
	\path[->] (preactrot) edge node[below]{$\sigma(\,\cdot\,+\beta)$} (featmaprot);
	\end{tikzpicture}
\end{center}

\subsection{Group-convolutional layers}

Given feature maps $\zeta^{(l)}(x,\theta)$, the group-convolutional layers perform an equivariant mapping of $\mathcal{R}_\alpha\zeta^{(l)}(x,\theta)$ to $\mathcal{R}_\alpha\zeta^{(l+1)}(x,\theta)$ under the group action $\mathcal{R}.$
The step of adding the bias and applying the activation function is equivariant by the same argument as in the first layer.
What is left to show is the equivariance $\left(\mathcal{R}_\alpha\zeta^{(l)}\circledast\Psi\right)(x,\theta) = \mathcal{R}_\alpha\left(\zeta^{(l)}\circledast\Psi\right)(x,\theta) = \mathcal{R}_\alpha y^{(l)}(x,\theta)$ of the group convolution.
Inserting a transformed feature map and writing the group convolution out explicitly yields:
\begin{align*}
& \left(\mathcal{R}_\alpha\zeta^{(l)}\circledast\Psi\right)(x,\theta) \\
=& \int\limits_{\mathbb{R}^2} \sum_{\phi\in\Theta} \zeta^{(l)}(\rho_{-\alpha}u,\phi-\alpha)\, \Psi(\rho_{-\phi}(x-u),\theta-\phi) \, \mathrm{d}u \,.
\end{align*}
Again, we substitute $\tilde{u}:=\rho_{-\alpha}u$ with $\left|\det\left(\frac{\partial \tilde{u}}{\partial u}\right)\right|=1.$
Furthermore, we let $\tilde{\phi}:=\phi-\alpha$ under which the sum is invariant thanks to the cyclic structure of the subgroup $\Theta,$ and we obtain
\begin{align*}
	& \left(\mathcal{R}_\alpha\zeta^{(l)}\circledast\Psi\right)(x,\theta) \\
	=& \int\limits_{\mathbb{R}^2} \sum_{\tilde{\phi}\in\Theta} \zeta^{(l)}(\tilde{u},\tilde{\phi})\, \Psi(\rho_{-\tilde{\phi}}(\rho_{-\alpha}x-\tilde{u}),(\theta-\alpha) - \tilde{\phi}) \, \mathrm{d}\tilde{u}\\ 
	=& \left(\zeta^{(l)}\circledast\Psi\right)(\rho_{-\alpha}x,\theta-\alpha) \\
	=& \rho_\alpha y^{(l+1)}(x,\theta-\alpha) \\
	=& \mathcal{R}_\alpha y^{(l+1)}(x,\theta)\,.
\end{align*}
This proves the equivariance of the intermediate layers.
Again, the relations are illustrated in a commutative diagram:
\begin{center}
	\begin{tikzpicture}
	\node (fmapin) {$\zeta^{(l)}(x,\theta)$}; 
	\node[below=1.5cm of fmapin] (fmapinrot) {$\rho_\alpha \zeta^{(l)}(x,\theta-\alpha)$}; 
	\node[right=4cm of fmapin] (preact) {$y^{(l+1)}(x,\theta)$}; 
	\node[below=1.5cm of preact] (preactrot) {$\rho_\alpha y^{(l+1)}(x,\theta-\alpha)$};
	
	\path[->] (fmapin) edge node[left]{$\mathcal{R}_\alpha$} (fmapinrot);
	\path[->] (fmapin) edge node[above]{$\circledast\Psi$} (preact);
	\path[->] (preact) edge node[right]{$\mathcal{R}_\alpha$} (preactrot);
	\path[->] (fmapinrot) edge node[below]{$\circledast\Psi$} (preactrot);
	\end{tikzpicture}
\end{center}

\subsection{Orientation max-pooling layer}

For rotation-invariant segmentation or classification we max-pool over orientations after the last group-convolutional layer.
The pooling step is itself equivariant and results in a rotated version of its output:
\begin{align*}
	\max_{\theta} \mathcal{R}_\alpha\zeta^{(l)}(x,\theta)\ 
	=\ & \max_{\theta} \rho_\alpha\zeta^{(l)}(x,\theta-\alpha) \\
	=\ & \rho_{\alpha} \left(\max_{\theta} \zeta^{(l)}(x,\theta-\alpha)\right) \\
	=\ & \rho_{\alpha} \left(\max_{\theta} \zeta^{(l)}(x,\theta)\right).
\end{align*}
The rotation operator commutes with the maximum over orientation channels because it acts on spatial coordinates only.
We again visualize the transformation behavior by a commutative diagram:
\begin{center}
	\begin{tikzpicture}
	\node (img) {$\zeta^{(l)}(x,\theta)$}; 
	\node[below=1.5cm of img] (imgrot) {$\rho_\alpha \zeta^{(l)}(x,\theta-\alpha)$}; 
	\node[right=4cm of img] (featmap) {$\max_\theta \zeta^{(l)}(x,\theta)$}; 
	\node[below=1.5cm of featmap] (featmapact) {$\rho_\alpha \max_\theta \zeta^{(l)}(x,\theta)$};
	
	\path[->] (img) edge node[left]{$\mathcal{R}_\alpha$} (imgrot);
	\path[->] (img) edge node[above]{$\max\limits_\theta$} (featmap);
	\path[->] (featmap) edge node[right]{$\rho_\alpha$} (featmapact);
	\path[->] (imgrot) edge node[below]{$\max\limits_\theta$} (featmapact);
	\end{tikzpicture}
\end{center}
In the case of classification the remaining spatial structure is pooled out such that the output is invariant under transformations of the input.

Instead of the maximum pooling which we applied in our experiments, one could also utilize average pooling layers.
The equivariance of average pooling can be derived in analogy to the derivation for maximum pooling.

\section{Derivation of the generalized He weight initialization scheme}\label{apx:HeWeightDeriv}

In this section we give the derivation of the generalized weight initialization scheme whose results are stated in the main paper.
For completeness we recall the assumptions going into the following calculations.
We consider the activation of a single neuron in layer $l,$
\begin{equation}\label{eq:Activation}
	\zeta_{\hat{c}x}^{(l)}=\max(0,y_{\hat{c}x}^{(l)}),
\end{equation}
where rectified linear units were chosen as nonlinearities.
The pre-nonlinearity activations are given by the convolution with filters $\Psi$ and summing over the input channels:
\begin{equation}\label{eq:preNonlinearityActivation}
\begin{split}
	y_{\hat{c}x}^{(l)}\ 
	&=\ \sum_{c}\left(\zeta_{c}^{(l-1)} \ast \Psi_{\hat{c}c}^{(l)}\right)_x + \beta_{\hat{c}}^{(l)} \\
	&=\ \sum_{c}\sum_{x'} \zeta_{c,x-x'}^{(l-1)} \Psi_{\hat{c}cx'}^{(l)} + \beta_{\hat{c}}^{(l)}.
\end{split}
\end{equation}
For convenience we shifted the addition of the bias to the pre-nonlinearity activations.
The filters are defined by
\[
	\Psi_{\hat{c}cx}^{(l)}\ =\ \sum_{q=1}^Q w_{\hat{c}cq}^{(l)}\psi_{qx} \,,
\]
that is, they are built from $Q$ real valued atomic filters $\psi_{q}$.
We keep the discussion general by not restricting the atomic filters to be steerable.
In analogy to \citet{Glorot10understandingthe} and \citet{DBLP:journals/corr/HeZR015} we assume the activations and gradients to be i.i.d.~and to be independent from the weights. We let the weights themselves be mutually independent and have zero mean but do \emph{not} restrict them to be identically distributed because of the inherent asymmetry coming from the different atomic filters. Furthermore we initialize all biases to be zero.

\subsection{Backpropagation}
In order to prevent vanishing or exploding gradients of the loss $\mathcal{E}$ due to inappropriate initialization we demand their variance $\Var\left[\frac{\partial\mathcal{E}}{\partial\zeta^{(l)}}\right]$ to be constant across all layers. 
It follows from \eqref{eq:Activation} and \eqref{eq:preNonlinearityActivation} that the gradient with respect to the activation $\zeta_{c_0 x_0}^{(l)}$ of a particular neuron in layer $l$ is given by
\begin{align}\label{eq:gradient}
	\frac{\partial\mathcal{E}}{\partial\zeta_{c_0x_0}^{(l)}}
	\ &=\ \sum\limits_{\hat{c},x} \frac{\partial\mathcal{E}}{\partial y_{\hat{c}x}^{(l+1)}} \frac{\partial y_{\hat{c}x}^{(l+1)}}{\partial\zeta_{c_0x_0}^{(l)}} \\\notag
	\ &=\ \sum\limits_{\hat{c},x} \frac{\partial\mathcal{E}}{\partial \zeta_{\hat{c}x}^{(l+1)}}\mathbb{I}_{y_{\hat{c}x}^{(l+1)}>0} \sum\limits_q w_{\hat{c}c_0q}^{(l+1)}\psi_{q,x-x_0},
\end{align}
where the indicator function $\mathbb{I}$ stems from the derivative of the rectified linear unit. Like \citet{DBLP:journals/corr/HeZR015} we assume the factors occurring in (\ref{eq:gradient}) to be statistically independent. Observing that $\E\left[w^{(l)}\right]$ and therefore also $\E\left[\frac{\partial\mathcal{E}}{\partial\zeta^{(l)}}\right]$ vanish, and without loss of generality setting $x_0=0$ this leads to
\allowdisplaybreaks
\begin{align*}
	&\ \Var\left[\frac{\partial\mathcal{E}}{\partial\zeta_{c_0x_0}^{(l)}}\right] 
	\ =\ \E\left[\left(\frac{\partial\mathcal{E}}{\partial\zeta_{c_0x_0}^{(l)}}\right)^2\right]\\
	=&\  \sum\limits_{\hat{c},\hat{c}'}\sum\limits_{x,x'}\sum\limits_{q,q'}\E\left[\frac{\partial\mathcal{E}}{\partial\zeta_{\hat{c}x}^{(l+1)}}\frac{\partial\mathcal{E}}{\partial\zeta_{\hat{c}'x'}^{(l+1)}}\right] \E\left[\mathbb{I}_{y_{\hat{c}x}^{(l+1)}>0}\mathbb{I}_{y_{\hat{c}'x'}^{(l+1)}>0}\right] \\
	&\qquad\qquad\qquad\qquad\qquad\ \cdot\ \E\left[w_{\hat{c}c_0q}^{(l+1)}w_{\hat{c}'c_0q'}^{(l+1)}\right]\psi_{q,x}\psi_{q',x'}\\
	=&\  \sum\limits_{\hat{c}}\sum\limits_{x}\sum\limits_{q}\E\left[\left(\frac{\partial\mathcal{E}}{\partial\zeta_{\hat{c}x}^{(l+1)}}\right)^2\right]\E\left[\mathbb{I}_{y_{\hat{c}x}^{(l+1)}>0}\right]\\
	&\qquad\qquad\qquad\qquad\qquad\qquad\quad\ \ \cdot\ \E\left[\left(w_{\hat{c}c_0q}^{(l+1)}\right)^2\right]\psi_{q,x}^2\\
	=&\ \sum\limits_{\hat{c}}\sum\limits_{x}\sum\limits_{q}\frac{1}{2}\Var\left[\frac{\partial\mathcal{E}}{\partial\zeta_{\hat{c}x}^{(l+1)}}\right]\Var\left[w_{\hat{c}c_0q}^{(l+1)}\right]\psi_{q,x}^2.
\end{align*}

The factor $\frac{1}{2}$ in the last line originates from the symmetric distribution of $y^{(l)}$ in conjunction with the indicator function.
Using the fact that the weights' variances are initialized to only depend on $q$ and the assumption of identically distributed gradients, both can be pulled out of the sums:
\begin{equation*}
\begin{split}
 &\ \Var\left[\frac{\partial\mathcal{E}}{\partial\zeta^{(l)}}\right]\\
=&\ \Var\left[\frac{\partial\mathcal{E}}{\partial\zeta^{(l+1)}}\right]\frac{\hat{C}}{2}\sum_q\Var\left[w_q^{(l+1)}\right]\,\norm{\psi_q}_2^2.
\end{split}
\end{equation*}
It seems reasonable to assign the contribution to the overall variance equally to the $Q$ summands.
Demanding the gradients' variances to be constant over layers then leads to the initialization condition
\begin{equation*}
\Var\left[w_q\right]=\frac{2}{\hat{C}Q\norm{\psi_q}_2^2}.
\end{equation*}

\subsection{Forward pass}
The calculation for the forward pass is similar to the case of backpropagation but considers the variance $\Var\left[y^{(l)}\right]$ of pre-nonlinearity activations instead of gradients.
As an exact calculation depends on the expectation value $\E\left[\zeta^{(l-1)}\right]$, which is not known, we approximate the result by exploiting the central limit theorem.
To this end, we note that the pre-nonlinearity activations \eqref{eq:preNonlinearityActivation} are summed up from \mbox{$C\sum_{q}|\supp{\psi_q}|$} independent terms of finite variance which is a relatively large number in typical networks.
This allows to approximate the variance by the asymptotic result implied by the central limit theorem:
\begin{align*}
	& \Var\left[y_{\hat{c}x}^{(l)}\right] \\
	\equalCLT\; & \Var\left[\sum\limits_{c}\sum\limits_{x'}\sum\limits_{q} \zeta_{c,x-x'}^{(l-1)} w_{\hat{c}cq}^{(l)}\psi_{qx'}\right]\\
	\stackrel{\text{ (CLT)} }{\approx}\; & \sum\limits_{c}\sum\limits_{x'}\sum\limits_{q} \Var\left[\zeta_{c,x-x'}^{(l-1)} w_{\hat{c}cq}^{(l)}\psi_{qx'}\right]\\
	\equalCLT\; & \sum\limits_{c}\sum\limits_{x'}\sum\limits_{q}\E\left[\left(\zeta_{c,x-x'}^{(l-1)}\right)^2\right]\E\left[\left(w_{\hat{c}cq}^{(l)}\right)^2\right]\psi_{qx'}^2\ .
\end{align*}
In the last step we made use of the independence of the weights from the previous layer's feature maps and \mbox{$\E[w]=0.$}
The symmetric distribution of weights leads to a symmetric distribution of pre-nonlinearity activations which in conjunction with ReLU nonlinearities implies \mbox{$\E[\zeta^2]=\frac{1}{2}\Var[y]$.}
To see this, note that the symmetry of the distribution of pre-nonlinearity activation leads on the one hand to
\begin{equation*}
\begin{split}
	\Var[y] \ &=\ \E[y^2]\\
	&=\ \int_\R \tilde{y}^2\, p_y(\tilde{y})\ \mathrm{d}\tilde{y}\\
	&=\ 2\int_{\R^+} \tilde{y}^2\, p_y(\tilde{y})\ \mathrm{d}\tilde{y}	\qquad\qquad\qquad\quad
\end{split}
\end{equation*}
and on the other hand to
\begin{align}
	\E[\zeta^2]
	\ =&\ \int_\R \tilde{\zeta}^2\, p_\zeta(\tilde{\zeta})\ \mathrm{d}\tilde{\zeta}\notag\\
	\ =&\ \int_\R \tilde{\zeta}^2\, \left(\frac{1}{2}\delta(0) + \Theta(\tilde{\zeta})p_y(\tilde{\zeta})\right)\ \mathrm{d}\tilde{\zeta}\notag\\
	\ =&\ \int_{\R^+} \tilde{\zeta}^2\, p_y(\tilde{\zeta})\ \mathrm{d}\tilde{\zeta}\,,\ \notag
\end{align}
where $\delta$ denotes the delta distribution and $\Theta$ is the Heaviside step function.
As before, we drop all indices which the random variables are independent from to compute the sums. This leads to
\begin{equation*}
\begin{split}
	\Var\left[y^{(l)}\right]\approx\Var\left[y^{(l-1)}\right]\frac{C}{2}\sum\limits_q\Var\left[w_q^{(l)}\right]\,\norm{\psi_q}_2^2,
\end{split}
\end{equation*}
which in turn suggests a weight initialization according to
\begin{equation*}\label{eq:initializationForwardprop}
\Var\left[w_q\right]=\frac{2}{CQ\norm{\psi_q}_2^2}
\end{equation*}
to ensure that the activations' variances are not amplified.

\vspace{3ex}
\subsection{Normalization of complex atomic filters}

The results derived above suggest to initialize the weights of each layer uniformly by
\begin{equation*}
	\Var\left[w_q\right]=\frac{2}{CQ\norm{\psi_q}_2^2} \quad \text{or} \quad \Var\left[w_q\right]=\frac{2}{\hat{C}Q\norm{\psi_q}_2^2}
\end{equation*}
after normalizing the atomic filters to $\norm{\psi_q}_2=1.$
An additional complication arises in our network construction where steerability is only preserved when the relative amplitude of the filters' real and imaginary parts is not changed.
While for circular harmonics both parts have equal norms in continuous space, this is not necessarily true for their sampled versions which rules out an independent normalization of the real and imaginary parts.
As a steerability consistent way of normalizing circular harmonics, we propose to adequately normalize their complex modulus.
The proper scale follows from $\norm{\psi}_2^2=\norm{\real\left[\psi\right]}_2^2+\norm{\imag\left[\psi\right]}_2^2$ for $\psi\in\mathbb{C}$ to be $\norm{\psi}_2=1$ for DC filters whose imaginary part vanishes and $\norm{\psi}_2=\sqrt{2}$ for non-DC filters.

\section{Details on the experimental setup}\label{apx:Experiments}

\begin{table}[t]
	\centering
	\small
	\begin{tabular}{lrr}
		\toprule 
		Operation & Filter Size & Feature Channels \\
		\toprule 
		Steerable input layer  & $7\times 7$ &  16 \\
		Steerable group convolution  & $5\times 5$ &  24 \\
		Spatial max pooling & $2 \times 2$ & \\
		\midrule
		Steerable group convolution  & $5\times 5$ &  32 \\
		Steerable group convolution  & $5\times 5$ &  32 \\
		Spatial max pooling & $2 \times 2$ & \\
		\midrule
		Steerable group convolution  & $5\times 5$ &  48 \\
		Steerable group convolution  & $5\times 5$ &  64 \\
		Global spatial pooling \\
		Global orientation pooling \\
		\midrule
		Fully connected  & & $64$ \\
		Fully connected  & & $64$ \\
		Fully connected + Softmax  & & $10$ \\
		\bottomrule
	\end{tabular}
	\caption{Architecture of the SFCNN used in the initial experiments on the resolution of sampled orientations and the rotational generalization.}
	\label{tab:architectureCNN1}
\end{table}

Here we give further details on the network architectures and the training setup of our experiments.

\subsection{Rotated MNIST}

For our initial experiments on the dependence on sampled orientations and the networks' rotational generalization capabilities we utilize the architecture given in Table~\ref{tab:architectureCNN1} as baseline.
Based on the results of these experiment we fix the number of sampled orientations to $\Lambda=16$ and tune the network architecture further.
We achieve the best benchmark results using the slightly larger network given in Table~\ref{tab:architectureCNN2}.
In particular, we found that increasing the size of the filter masks improved the results.
Both architectures consist of one steerable input layer which maps the input images to the group, five following group convolutional layers and three fully connected layers.
After every two steerable filter layers we perform a spatial $2\times2$ max-pooling.
The orientation dimension and the remaining spatial dimensions are pooled out globally after the last convolutional layer.
We normalize the activations by adding batch normalization layers \citep{ioffe2015batch} after each convolutional and fully connected layer.
The batch normalization on the group does not interfere with the equivariance when the responses are normalized by averaging over both spatial and orientation dimensions.

The number of feature channels stated in the tables refers to the number of learned filters $\hat{C}$ of the corresponding layer.
As these filters are themselves applied with respect to $\Lambda$ orientations we end up with $\hat{C}\Lambda$ responses; e.g. $24\cdot16=384$ effective responses in the first layer of the smaller network.
Note that the extraction of this comparatively large number of responses without overfitting is possible because the rotational weight sharing leads to an increased parameter utilization (in the sense of \citet{cohen2016steerable}) by a factor $\Lambda$.

All networks are trained for $40$ epochs using the Adam optimizer \citep{kingma2015adam} with standard parameters.
The initial learning rate is set to $0.015$ and is decayed exponentially with a rate of $0.8$ per epoch starting from epoch~$15.$
We regularize the weights with an elastic net penalty with hyperparameters $\lambda_{L1}=\lambda_{L2}$ which are set to $10^{-7}$ and $10^{-8}$ for the convolutional and fully connected layers respectively.
Dropout~\citep{srivastava2014dropout} is used only in the fully connected layers with a dropping probability of $p=0.3$.

\begin{table}[t]
	\centering
	\small
	\begin{tabular}{lrr}
		\toprule 
		Operation & Filter Size & Feature Channels \\
		\toprule 
		Steerable input layer  & $9\times 9$ &  24 \\
		Steerable group convolution  & $7\times 7$ &  32 \\
		Spatial max pooling & $2 \times 2$ & \\
		\midrule
		Steerable group convolution  & $7\times 7$ &  36 \\
		Steerable group convolution  & $7\times 7$ &  36 \\
		Spatial max pooling & $2 \times 2$ & \\
		\midrule
		Steerable group convolution  & $7\times 7$ &  64 \\
		Steerable group convolution  & $5\times 5$ &  96 \\
		Global spatial pooling \\
		Global orientation pooling \\
		\midrule
		Fully connected  & & $96$ \\
		Fully connected  & & $96$ \\
		Fully connected + Softmax  & & $10$ \\
		\bottomrule
	\end{tabular}
	\caption{Architecture of the SFCNN used with $\Lambda=16$ sampled orientations in the final benchmarking experiments on rotated MNIST.}
	\label{tab:architectureCNN2}
\end{table}

\begin{figure*}
	\centering
	\includegraphics[width=.72\textwidth]{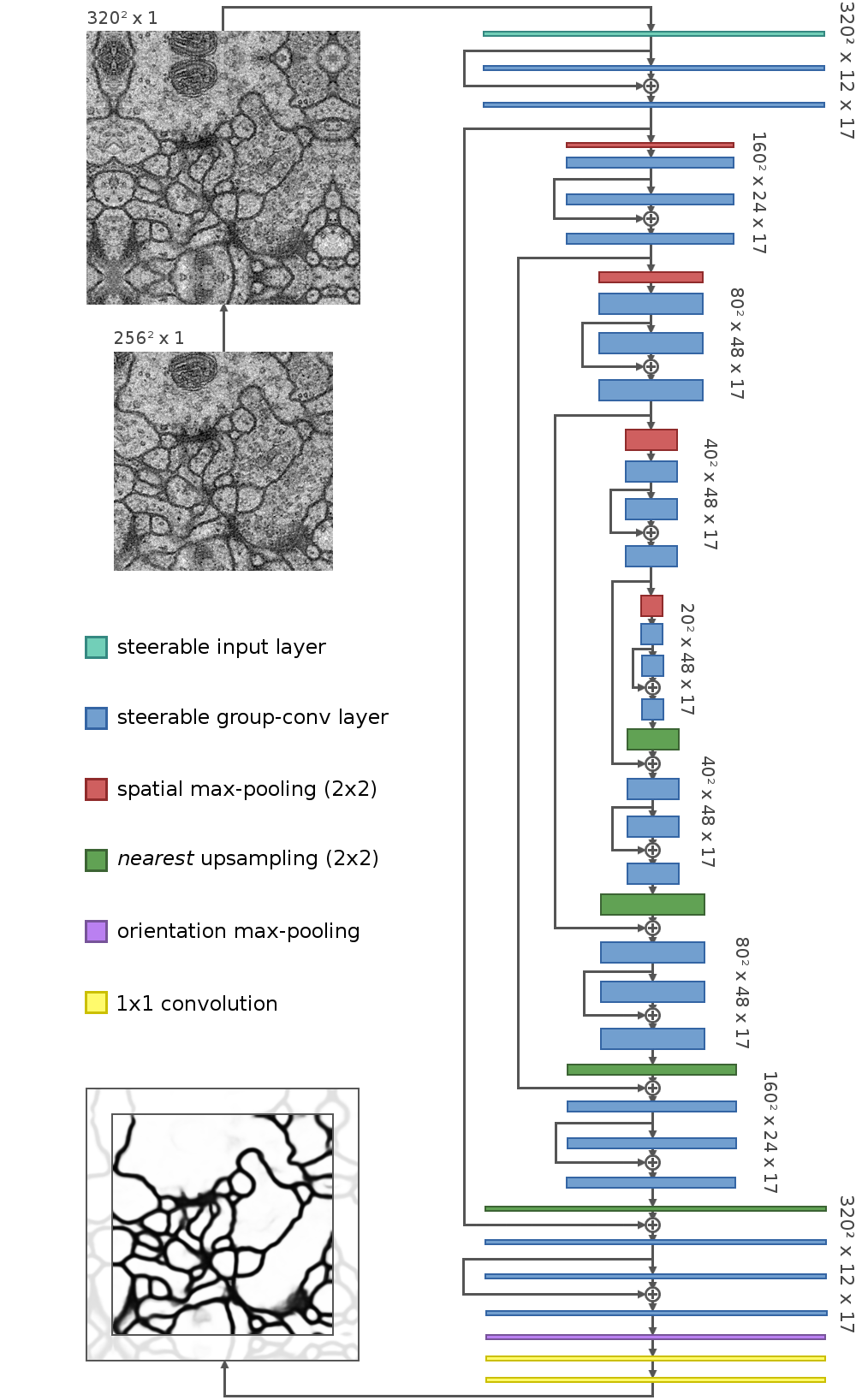}
	\caption{
		Network architecture used to predict the membrane probability map for the ISBI 2012 EM segmentation challenge.
		The topology is inspired by the U-Net \cite{ronneberger2015u} and FusionNet \cite{quan2016fusionnet} but uses the proposed steerable group-convolution layers with $\Lambda=17$ orientations.
		To mitigate boundary artifacts we feed reflect-padded images into the network.
	}
	\label{fig:unet_architecture}
\end{figure*}

\subsection{ISBI 2012 EM segmentation challenge}

The network architecture used to segment the membranes from raw EM images of neural tissue for the ISBI EM segmentation challenge is visualized in Figure~\ref{fig:unet_architecture}.
Inspired by the U-Net~\cite{ronneberger2015u} it is build as a symmetric encoder-decoder network with additional skip-connections between stages of the same resolution.
This allows to extract semantic information from a large field of view while at the same time preserving precise spatial localization.
Further, we adopt two modifications from \cite{quan2016fusionnet}: we do not concatenate the skipped feature maps but add it to the decoder features upsampled from the previous stage, and we use intermediate residual blocks (here of depth 1).
On the highest resolution level we learn $\hat{C}=12$ filters, applied in $\Lambda=17$ orientations which corresponds to $\hat{C}\Lambda=204$ effective channels.
The number of filters is doubled when going to the second and third level and is afterwards kept constant since we did not observe further gains in performance when adding more channels.
All group-convolutional layers utilize kernels of size $7\times7$ pixels while the input layer applies $11\times11$ pixel kernels.

As input, we feed the network cropped regions of $256\times256$ pixels which are padded to $320\times320$ pixels by reflecting a region of $32$ pixels around the borders to alleviate boundary artifacts.
The padded regions are augmented by random elastic deformations, reflections and rotations by multiples of $\frac{\pi}{2}$.
After the decoder we max-pool over orientations to obtain locally invariant features and crop out $256\times256$ pixels centrally.
Two subsequent $1\times1$ convolution layers map these features pixel-wise to the desired probability map.

The network is optimized by minimizing the spatially averaged binary cross-entropy loss between predictions and the ground truth segmentation masks using the ADAM optimizer.
As on the rotated MNIST dataset we regularize the convolutional weights with an elastic net penalty with hyperparameters $\lambda_{L1}=\lambda_{L2}$ set to $10^{-7}$ and $10^{-8}$ for the steerable and $1\times1$ convolution layers respectively.
Here we chose a dropout probability of $p=0.4$ both in the steerable as well as in the $1\times1$ convolution layers.
The learning rate is decayed exponentially by a factor of $0.85$ per epoch starting from an initial rate of $5\cdot10^{-2}$.

\clearpage
{\small
\bibliographystyle{IEEEtranN}
\bibliography{steerableNetworkCVPR2018}
}

\end{document}